\documentclass[runningheads]{llncs}
\usepackage{graphicx}

\usepackage{tikz}
\usepackage{comment}
\usepackage{amsmath,amssymb} 
\usepackage{color}

\usepackage[normalem]{ulem}
\usepackage{multirow,booktabs}
\usepackage{footnote}
\usepackage{kotex}
\usepackage{array}
\usepackage[flushleft]{threeparttable}
\usepackage{wrapfig}

\usepackage{enumitem}
\usepackage{bm}

\newcommand{\etal}{\textit{et al.}}

\definecolor{Highlight}{HTML}{39b54a}  

\newcommand{\PAR}[1]{\vskip4pt \noindent {\bf #1.~}} 

\usepackage{cite}
\usepackage[pagebackref,breaklinks,colorlinks,bookmarks=false]{hyperref}

\usepackage[accsupp]{axessibility}  

\begin{document}
\pagestyle{headings}
\mainmatter
\def\ECCVSubNumber{1946}  

\title{COO: Comic Onomatopoeia Dataset for Recognizing Arbitrary or Truncated Texts}


\titlerunning{COO for Recognizing Arbitrary or Truncated Texts}
%
\author{Jeonghun Baek \and Yusuke Matsui \and Kiyoharu Aizawa}
\authorrunning{J. Baek et al.}
%
\institute{The University of Tokyo\\
\email{\{baek,matsui,aizawa\}@hal.t.u-tokyo.ac.jp}}
\maketitle

\begin{abstract}
Recognizing irregular texts has been a challenging topic in text recognition.
To encourage research on this topic, we provide a novel comic onomatopoeia dataset (COO), which consists of onomatopoeia texts in Japanese comics.
COO has many arbitrary texts, such as extremely curved, partially shrunk texts, or arbitrarily placed texts.
Furthermore, some texts are separated into several parts.
Each part is a truncated text and is not meaningful by itself.
These parts should be linked to represent the intended meaning.
Thus, we propose a novel task that predicts the link between truncated texts.
We conduct three tasks to detect the onomatopoeia region and capture its intended meaning: text detection, text recognition, and link prediction.
Through extensive experiments, we analyze the characteristics of the COO.
Our data and code are available at \url{https://github.com/ku21fan/COO-Comic-Onomatopoeia}.
\end{abstract}

\section{Introduction}\label{sec:intro}
Along with the development of deep neural networks, text recognition methods have significantly improved. 
Currently, most state-of-the-art methods can easily recognize simple horizontal texts.
Recently, the research trend has progressed to recognize more irregular texts: recognizing horizontal text to recognizing arbitrary-shaped text such as curved or perspective text in scene images~\cite{RRPN,textsnake,PSENet,PAN,Boundary,Contour,DRRG,MTSv3,ABCNetv2,PCR,FCENet}.
We expect that \textit{studies on more irregular texts will further improve the text recognition methods}.

\begin{figure}[t]
\centering
    \includegraphics[width=1.0\textwidth]{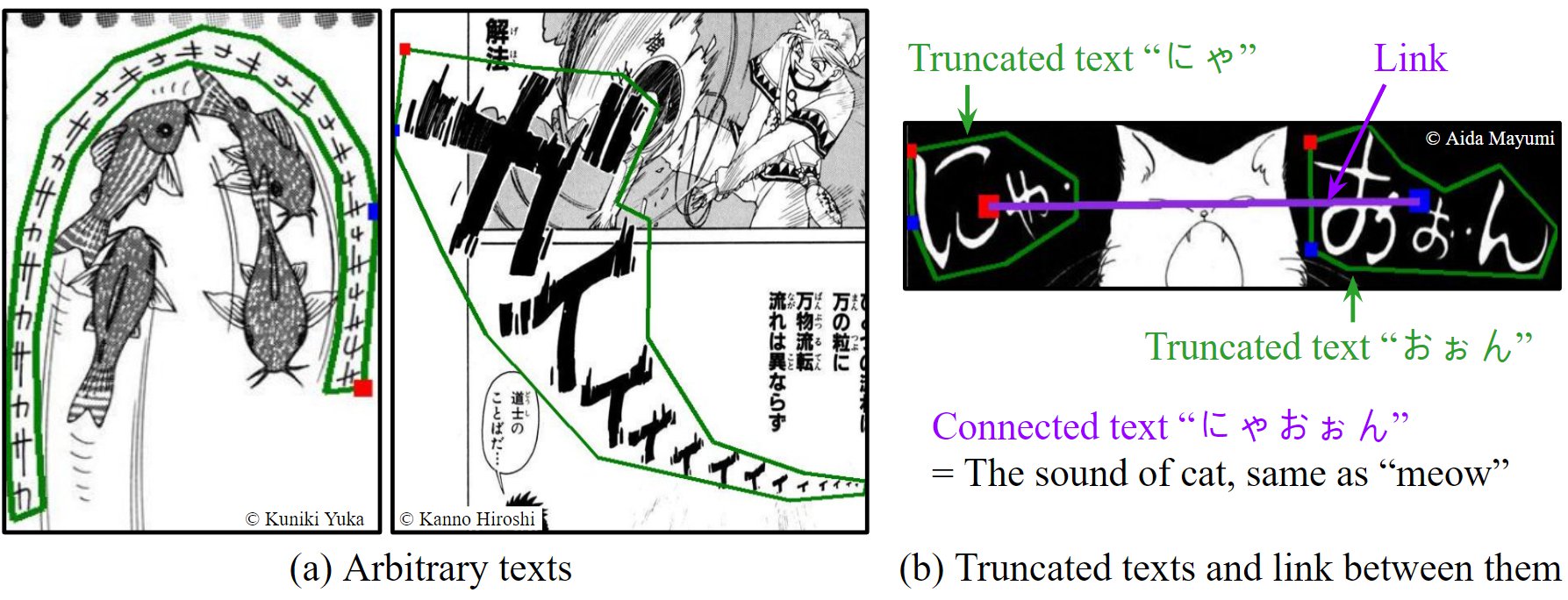}
    \vspace{-7mm}
    \caption{Visualization of comic onomatopoeia dataset \textit{COO}. 
    Red and blue squares denote the start and end points of each annotation, respectively.
    The purple line denotes the link between truncated texts}
    \label{fig:teaser}
\end{figure}

To encourage these studies, we provide a novel comic onomatopoeia dataset (COO), which contains more irregular texts.
After investigating various text datasets from English~\cite{IC13,CUTE80,IC15,COCO,Uber,totaltext,TextOCR,Comic-2017amazing} to other languages~\cite{RCTW,ArT,LSVT,ReCTS,MLT19,eBDtheque2013,manga109,manga109_2}, we find that onomatopoeia texts in the Japanese comic dataset~(Manga10-9~\cite{manga109}) have arbitrary shapes or are arbitrarily placed in the image.
Onomatopoei-as are written texts that represent the sound or state of objects (humans, animals, and so on).
To exaggerate the sound or state of the object, onomatopoeias are typically written in irregular shapes or placed at unexpected positions.
Fig.~\ref{fig:teaser}~(a) illustrates examples of COO: 
(left) shows extremely curved text, (right) shows partially shrunk text, and part of the text is on the object.

Onomatopoeia in Japanese comics is sometimes separated into several parts, as shown in Fig.~\ref{fig:teaser}~(b).
When separated, each part is a truncated text. 
Each truncated text does not fully represent the meaning.
After truncated texts are connected, the connected text represents the intended meaning.
For example, the truncated texts ``にゃ'' and ``おぉん'' in Fig.~\ref{fig:teaser}~(b) do not represent the meaning independently.
When they are connected into ``にゃおぉん'', the connected text represents the meaning: the sound of cat, same as ``meow''.
To correctly capture the meaning of truncated texts, \textit{we propose a novel task that predicts the link between truncated texts}.
By using the link information, we connect truncated texts to capture the intended meaning.
To solve this task, we formulate the task as the sequence-to-sequence problem~\cite{seq2seq}, and propose a model named M4C-COO, a variant of multimodal multi-copy mesh (M4C)~\cite{M4C}.

Considering truncated texts, we conduct three tasks to detect the onomatop-oeia region and capture its intended meaning:
1) Text detection: The model takes an image, and outputs the regions of onomatopoeias. 
2) Text recognition: The model takes the region of onomatopoeia, and outputs the text in the region.
3) Link prediction: The model takes the regions and texts of onomatopoeias, and outputs the links between truncated texts.
With extensive experiments using state-of-the-art methods, we analyze the characteristics of COO and the limitation of current models.
We hope that these analyses inspire and encourage future work on recognizing arbitrary or truncated texts.

Among three tasks, we mainly focus on text recognition and link prediction.
Because they are somewhat different from existing tasks, they can hinder using our dataset.
To prevent it, we provide decent baselines for them.
Traditional text recognition task generally recognizes horizontal or curved texts.
However, the COO has many vertically long texts: In 72.5\% of onomatopoeia regions, the height is greater than the width.
To address vertically long texts, we introduce several effective techniques.
In the case of the link prediction task, it is a novel task, and thus we introduce it thoroughly.

\clearpage

In summary, our main contributions are as follows:
\begin{itemize}[leftmargin=*]
    \item We construct a novel challenging dataset \textit{COO} to encourage the research on recognizing arbitrary or truncated texts.
    \item COO has many vertically long texts, and we investigate several techniques that are effective to recognize vertically long texts.
    \item COO has some truncated texts, and they should be linked.
    We propose a novel task, which predicts the link between truncated texts, and a M4C-COO model for this task.
\end{itemize}

\section{COO: Comic Onomatopoeia Dataset}\label{sec:Comic}
Most onomatopoeias in COO are arbitrary-shaped or arbitrarily placed. 
In addition, they are written in informal fonts or various sizes.
This section introduces the visualization, annotation guideline, statistics, and analysis of our dataset.
More details are presented in the supplementary materials.

\begin{figure}[t]
\centering
     \includegraphics[width=\linewidth]{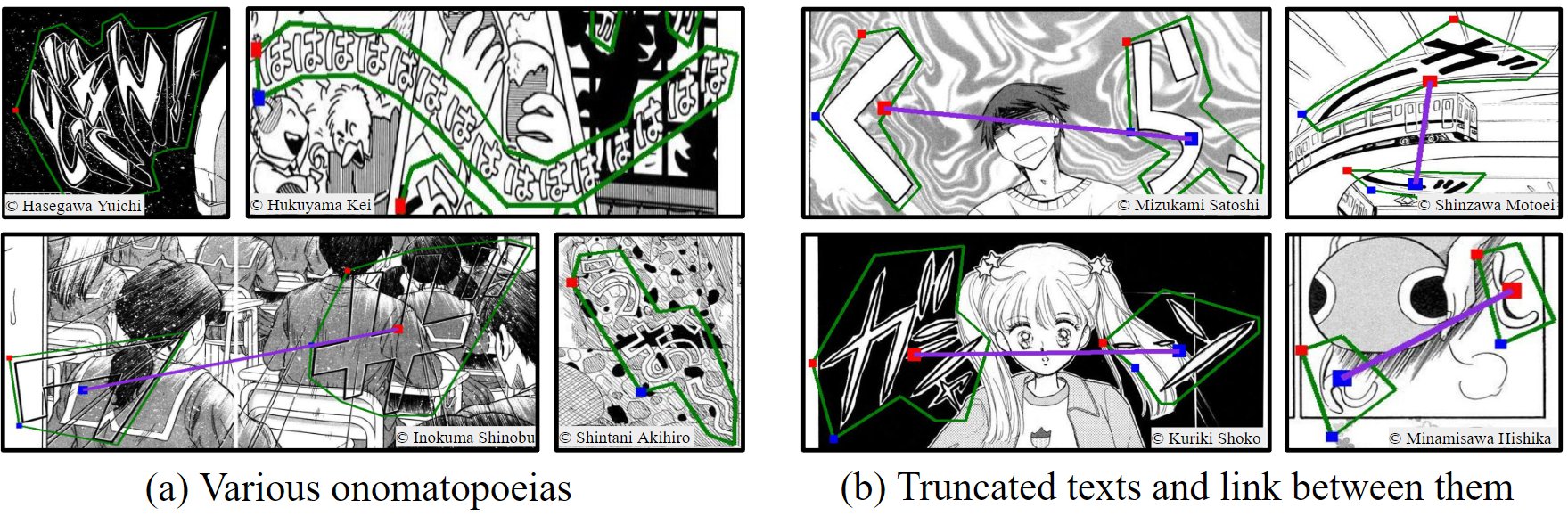}
    \vspace{-7mm}
    \caption{Visualizations of COO.
    Each example shows diversity of onomatopoeias}
    \label{fig:data}
\end{figure}

\subsection{Why Use Onomatopoeias of Japanese Comics?}
We use onomatopoeias of Japanese comics rather than English comics.
Because of following three reasons: 

\begin{enumerate}[leftmargin=*]
    \item Japanese comics have various types of onomatopoeias.
    Fig.~\ref{fig:teaser} and Fig.~\ref{fig:data} show many arbitrary or truncated texts. 
    As arbitrary texts, Fig.~\ref{fig:data}~(a)~(top row) shows three-dimensional or curved texts.
    Fig.~\ref{fig:data}~(a)~(bottom row) shows transparent texts on the objects that look similar to background objects.
    Fig.~\ref{fig:data}~(b) shows truncated texts.
    
    \item Japanese comics have more onomatopoeias than English comics. 
    We compared Japanese comic dataset Manga109~\cite{manga109}, and English comic dataset COMICS~\cite{Comic-2017amazing}. 
    Manga109 has \textit{5.8 onomatopoeias per page on average},
    whereas COMICS has much fewer (492 onomatopoeias in the first 5,000 images).
    
    \item Japanese language has more diverse onomatopoeias than most languages. According to Petersen~\cite{comic_studies}, ``The reason sounds in manga are so rich and varied is also in part due to the nature of the Japanese language that has a much wider range of onomatopoeic expressions than most languages.''
    \vspace{-2mm}
\end{enumerate}

\begin{wrapfigure}{r}{0.5\textwidth}
\centering
    \vspace{-8mm}
    \includegraphics[width=0.48\textwidth]{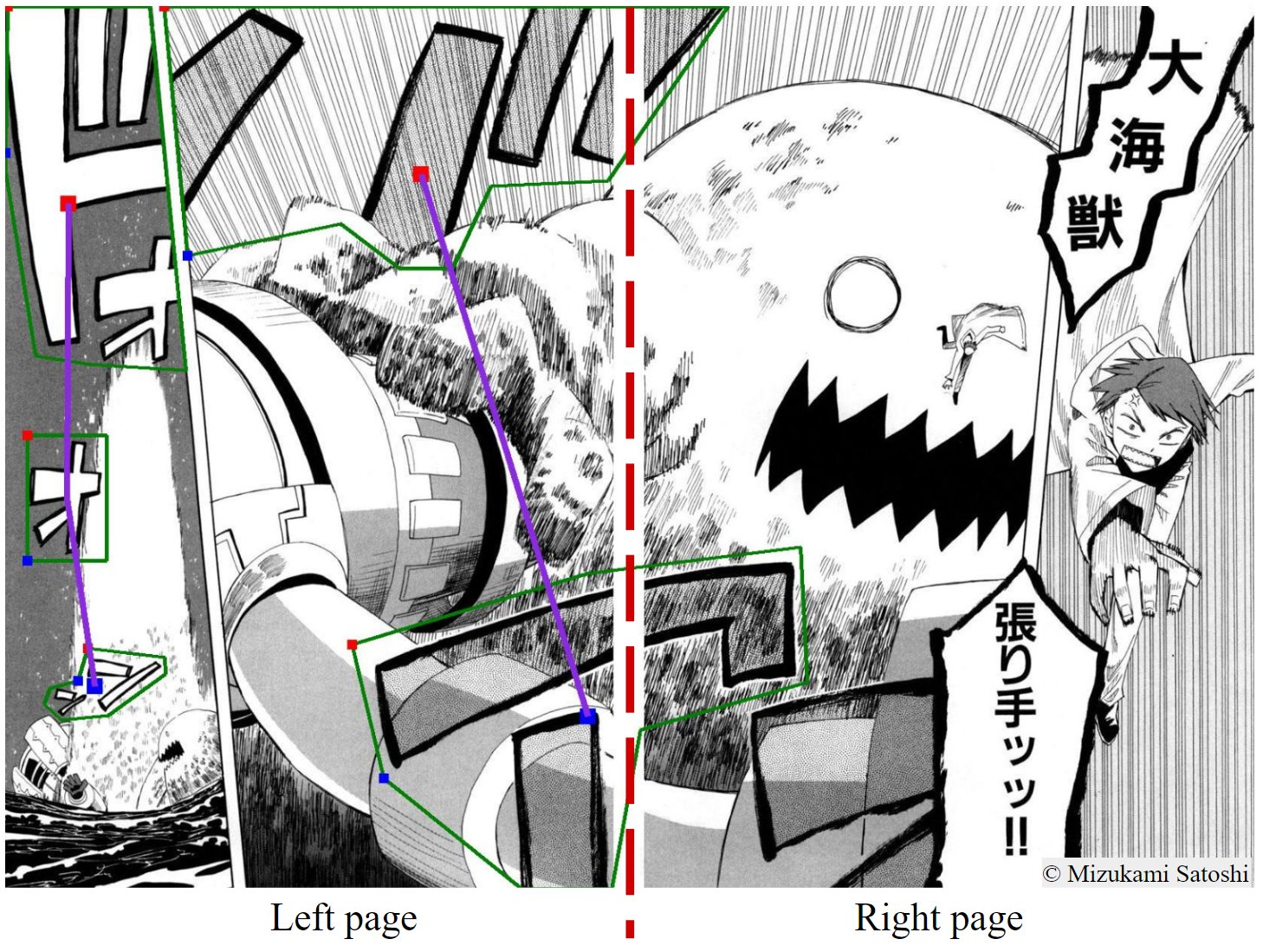}
    \vspace{-4mm}
    \caption{Each image in Manga109 consists of two pages, and it is used as input data for text detection task}
    \label{fig:twopage}
    \vspace{-10mm}
\end{wrapfigure}

\subsection{Label Annotation}
For each comic onomatopoeia, we create annotation data according to three tasks: 
1) Text detection: Annotate polygon regions. 
2) Text recognition: Annotate texts. 
3) Link prediction: Annotate links between truncated texts.

We annotate comic onomatopoeias in Manga109~\cite{manga109}.
Manga109 consists of 109 Japanese comics.
Each image in Manga109 consists of two pages (left and right pages) because some objects or onomatopoeias lie across two pages, as shown in Fig.~\ref{fig:twopage}. 

\PAR{Polygon regions of onomatopoeia}
We use polygon annotations instead of bounding box annotations to minimize regions irrelevant with texts.
We place the points that represent the contour of the onomatopoeia.
The points are placed clockwise starting from the top left of the onomatopoeia and ending with the bottom left.
Red and blue squares in Fig.~\ref{fig:data} denote start and end points of each annotation, respectively.
Because most state-of-the-art methods are developed with single-line annotations, we split a multi-lined onomatopoeia into single lines as much as possible.

\PAR{Texts of onomatopoeia}
While the Japanese language consists of Hiragana (e.g. ``あ'', ``い''), Katakana (e.g. ``ア'', ``イ''), and Chinese characters (e.g. ``任'', ``意''), Japanese comic onomatopoeias are typically written in Hiragana or Kata-kana.
Thus, Chinese characters are not the target of annotation.
We annotate Hiragana, Katakana, and some special symbols for each onomatopoeia.
Generally, most comic onomatopoeias are written in informal fonts or by drawing.

\PAR{Link between truncated texts}
Link annotation is conducted on polygon and text annotations.
The link between truncated texts is determined by the meaning of onomatopoeia.
Onomatopoeias have a link if the following conditions are satisfied:
1) Two or more onomatopoeias are separated in the image.
2) By themselves, they do not fully represent the intended meaning.
3) When they are connected, they represent the intended meaning.
The purple line in Fig.~\ref{fig:data} denotes the link between truncated texts.

The annotation was performed with an annotation team consisting of 15 annotators and 3 annotation checkers.
After all onomatopoeias in 109 comics were annotated, annotation checkers performed the initial check.
After that, we (authors) checked the annotations over three times and provided feedback for re-annotation to ensure annotation quality.
As a result, annotations have been revised over three times.

\clearpage

\begin{wraptable}{r}{0.6\textwidth}
  \vspace{-11mm}
  \caption{Statistics on the COO dataset}
  \tabcolsep=0.12cm
    \begin{center}
        \begin{tabular}{@{}lrrrr@{}}
            \toprule
            Count type & Total & Train & Valid & Test \\
            \midrule
            Images & 10,602 & 8,763 & 890 & 949 \\
            Comic volumes & 109 & 89 & 10 & 10 \\ 
            \midrule
            Polygon & 61,465 & 50,064 & 4,636 & 6,765 \\
            Link & 2,261 & 1,923 & 161 & 177 \\
            \midrule
            Vocabularies & 13,272 & 11,635 & 1,915 & 2,251 \\
            Character types & 182 & 182 & 163 & 166 \\
            \bottomrule
        \end{tabular}
    \label{tab:stat}
    \end{center}
    \vspace{-10mm}
\end{wraptable}

\subsection{Dataset Analysis} \label{sec:sub_data_analysis}
Table~\ref{tab:stat} presents the statistics of COO dataset.
COO has 61,465 polygons in total. 
If we regard the polygon that has more than 4 points as curved, 
the ratios of the curved, quadrilateral, and rectangular annotations are 61.3\%, 15.4\%, and 23.4\%, respectively.
The average number of points on all polygons is 6.3.
COO has 2,261 links in total, and one link appears every five pages on average.
Most links are between two truncated texts.
The numbers of links made by three, four, and five truncated texts are 132, 11, and 1, respectively.

\begin{wrapfigure}{r}{0.6\textwidth}
\centering
    \vspace{-8mm}
    \includegraphics[width=0.58\textwidth]{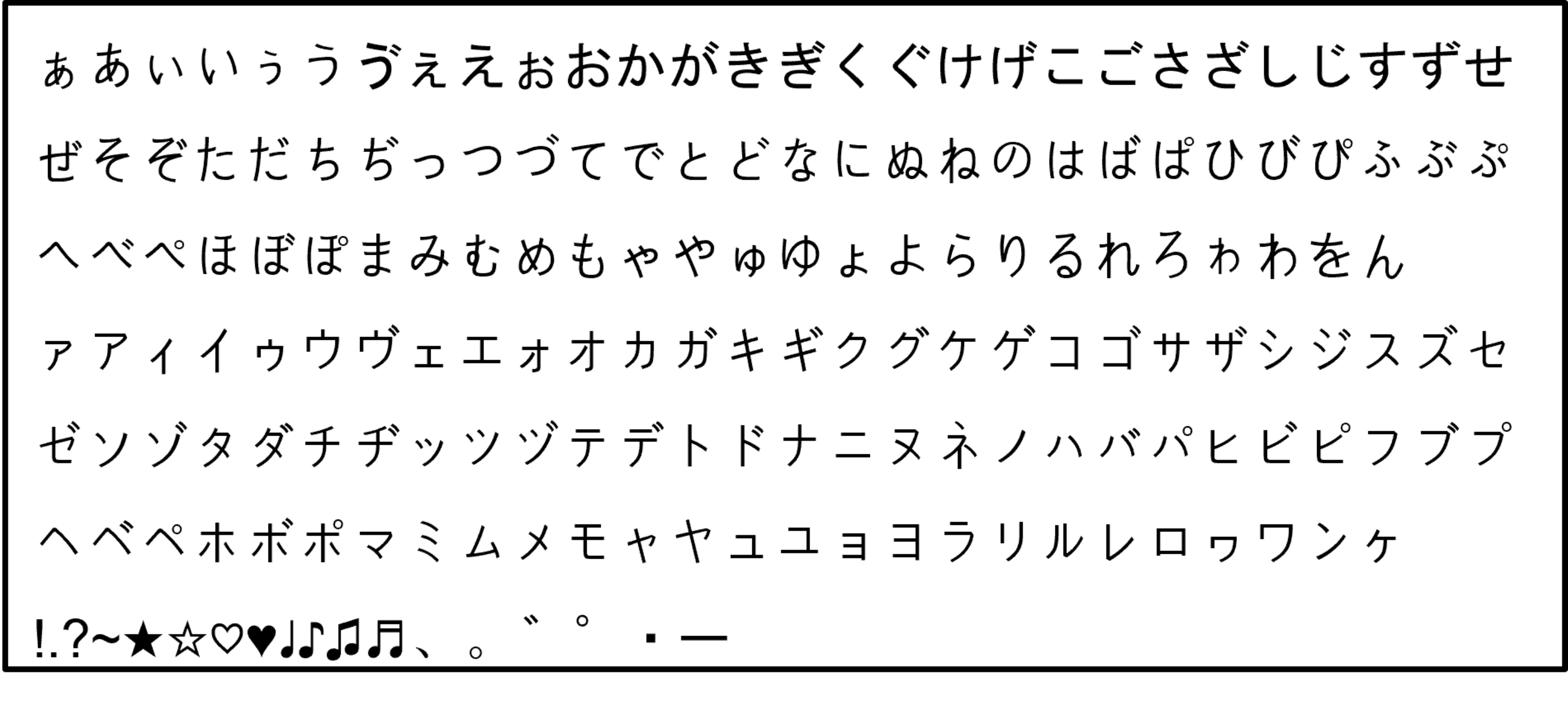}
    \vspace{-4mm}
    \caption{Total 182 character types of COO. 
    Because COO does not contain Chinese characters, the number of character types is much smaller than that in the Japanese language (over 2,000 characters)}
    \label{fig:char}
    \vspace{-6mm}
\end{wrapfigure}

The number of character types is 182, and Fig.~\ref{fig:char} shows all of them.
To recognize English texts, one may use 94 characters, including alphanumerics and symbols, as done in ASTER~\cite{ASTER}.
To recognize general Japanese texts, one should use thousands of Chinese characters.
Then, the number of character types exceeds thousands, and the increment of the character types makes text recognition difficult. 
However, comic onomatopoeias do not include thousands of Chinese characters, and character types do not increment considerably, only 94 to 182.
This indicates that the difficulty from the increment of the character types is little, and we can focus on recognizing arbitrary or truncated texts.

COO can be also used for comic analysis or comic translation.
For example, 79\% of links start from the left and end with the right.
This is an interesting characteristic because Japanese comics generally read right to left while the order of most truncated texts is reverse.
For other example, in our manual check over each truncated texts, we find that an object is typically placed between truncated texts.
This is assumed to be a drawing technique to represent the sound or state of the object dramatically.

\subsection{Comparison with Existing Arbitrary Scene Text Datasets}
Comic onomatopoeias exhibit various shapes and sizes, and are placed arbitrarily in the image.
They are close to scene text rather than document text.
There are several existing arbitrary-shaped scene text datasets: CUTE~\cite{CUTE80}, CTW1500~\cite{CTW1500}, Total-Text~\cite{totaltext}, ArT~\cite{ArT}, and TextOCR~\cite{TextOCR}.
They have polygon annotations for curved texts. 
While these datasets focus on arbitrary-shaped texts, our dataset COO focuses on both arbitrary-shaped texts and arbitrarily placed texts.
Furthermore, while most texts in other datasets are horizontal or curved, and are not separated into several parts, our dataset has many vertical texts or texts separated into several parts (truncated texts).
Thus, we mainly focus on vertical text recognition and link prediction between truncated texts.

Our dataset contains only Japanese comic onomatopoeias whereas other datasets mainly contain English or Chinese texts.
One may concern that methods developed on our dataset may not be generalized to other cases.
However, we believe that the algorithm developed on Japanese comic onomatopoeias can be generalized to other cases because the algorithm developed on the small English benchmark data was generalized to other languages.
Scene text detection and recognition methods have been developed based on English datasets.
According to the benchmark paper of scene text recognition~\cite{TRBA}, the total number of English benchmark evaluation data is 8,539. 
The number of character types and vocabulary of the data are 79 and 3,940, respectively. 
The data definitely do not cover all the texts in real life. 
However, the developed algorithm for competing on this small English benchmark data did not differ from the winning algorithms in ICDAR2019 competitions to recognize multi-lingual texts~\cite{MLT19} and Chinese texts~\cite{ArT,LSVT,ReCTS}.

\begin{figure}[t]
    \begin{center}
    \includegraphics[width=\linewidth]{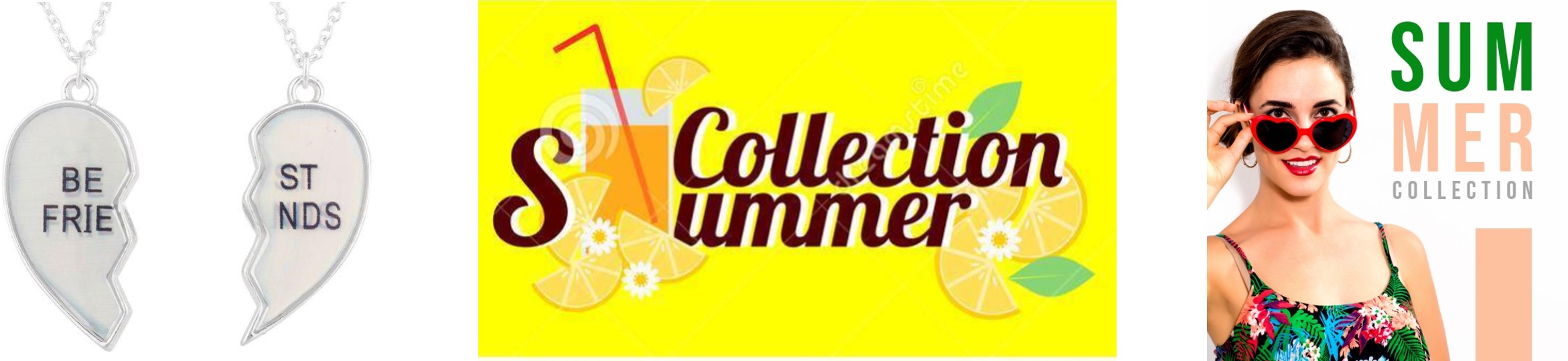}
    \end{center}
    \vspace{-6mm}
    \caption{Examples of truncated texts in English}
    \vspace{-4mm}
    \label{fig:sep}
\end{figure}

\subsection{Truncated Texts in English}
Truncated texts (onomatopoeias) in Japanese comics and link prediction for connecting them might be considered too special.
However, it is not, and similar problems also occur in other cases.
Fig.~\ref{fig:sep}\footnote{The images in Fig.~\ref{fig:sep} can be found here: 
\href{https://bit.ly/SW--4}{left}, 
\href{https://bit.ly/SW--2}{middle}, 
\href{https://bit.ly/SW--3}{right} (accessed 03-08-2022)}~(left) shows that the words on necklace are separated into two pieces: ``BEST'' $\rightarrow$ ``BE'' and ``ST'', and ``FRIENDS'' $\rightarrow$ ``FRIE'' and ``NDS''. 
(middle) shows that the word ``Summer'' is separated into ``S'' and ``ummer''. 
(right) shows that the word ``SUMMER'' is separated into ``SUM'' and ``MER''. 
Like the (right) image, we sometimes see a word separated in multiline in the poster or commercials.
In general, current state-of-the-art methods are specialized in single-line recognition, not multiline.
Here, we can use link prediction to connect them (``SUM'' and ``MER'') and capture the intended meaning (``SUMMER''). 
Extending the link prediction to these cases can be a new problem of our community.

\section{Methods for Three Tasks}\label{sec:method}
We summarize our methods for three tasks. 
Text detection and recognition are well-known tasks; thus, we skip details of model description.
Meanwhile, since link prediction between truncated texts is a novel task proposed in this study, we thoroughly introduce the M4C-COO model for the task.
More details are in the supplementary materials.

\subsection{Text Detection}
Text detection methods are mainly categorized into regression-based~\cite{RRPN,Boundary,Contour,DRRG,ABCNetv2,PCR,FCENet} and segmentation-based methods~\cite{textsnake,PSENet,PAN,MTSv3}.
To investigate the appropriate approach for comic onomatopoeia, we use two methods in each category.
Specifically, we use ABCNet~v2~\cite{ABCNetv2} and MTS~v3~\cite{MTSv3} as representatives of regression-based and segmentation-based methods, respectively.
Both methods were originally proposed for the text spotting task in which text detection and recognition tasks are combined.
However, these methods also provided results of using only the text detection part and showed state-of-the-art performance.
We take the only text detection part and use them as text detectors. 
Furthermore, MTS~v3 exhibits superior performance for rotated text detection. 
We expect that MTS~v3 can also detect vertical texts in our dataset.

\subsection{Text Recognition}
In this study, the well-known model called TPS-ResNet-BiLSTM-Attention (TR-BA)~\cite{TRBA} is used.
TRBA is created by combining existing methods such as RARE \cite{RARE} and FAN~\cite{FAN}.
TRBA takes four steps to recognize texts:
1) Rectify input image with TPS transformation~\cite{bookstein1989principal}. 
2) Convert rectified images into visual features by ResNet~\cite{ResNet}. 
3) Convert visual features into contextual features by BiLSTM. 
4) From contextual features, predict character string with attention module~\cite{NMT-attn}.

\subsection{Link Prediction}
In this study, we formulate the link prediction task into the sequence-to-sequence problem~\cite{seq2seq}.
The model takes the sequence of all onomatopoeias in an image, and outputs the sequence of truncated texts.
The sequence of truncated texts consists of pairs of truncated texts and the delimiter symbol $<$d$>$ which divides pairs of truncated texts.
Under this setting, predicting links between truncated texts is the same as predicting the sequence of truncated texts from the input sequence.

\begin{figure}[t]
\centering
    \includegraphics[width=\linewidth]{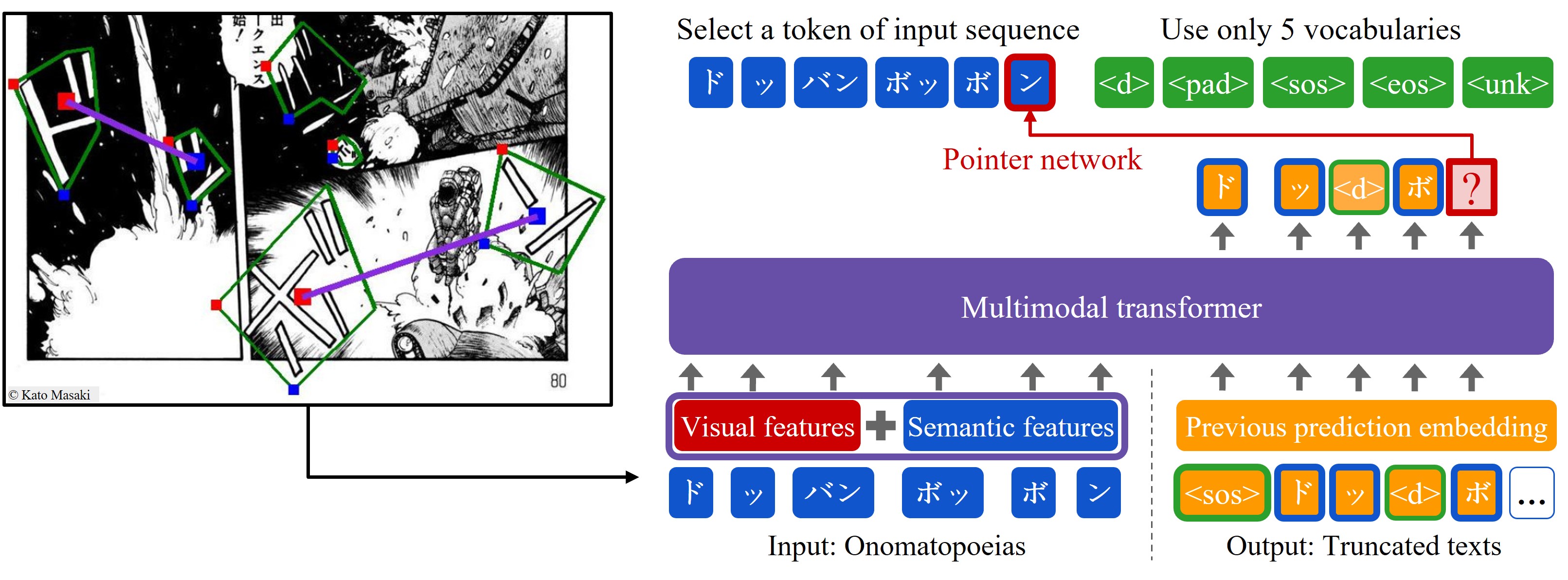}
    \vspace{-4mm}
    \caption{M4C-COO takes the sequence of all onomatopoeias in an image and outputs the sequence of truncated texts
    }
    \label{fig:M4C-COO}
\end{figure}

An example of input and output sequences is as follows.
Given the input sequence as below and truncated texts are (1) ``ド'' and ``ッ'' separated from ``ドッ'', and (2) ``ボ'' and ``ン'' separated from ``ボン'' (when two pairs of truncated texts exist), the output sequence is as follows.
\begin{itemize}[label={}]
    \item Input sequence: [ド, ッ, バン, ボッ, ボ, ン] 
    \item Output sequence: [ド, ッ, $<$d$>$, ボ, ン]
\end{itemize}
By dividing the output sequence with $<$d$>$, we obtain two lists [ド, ッ] and [ボ, ン]. 
By connecting each of the lists, we obtain connected texts ``ドッ'' and ``ボン''.
Fig.~\ref{fig:M4C-COO} illustrates this example.

To solve this sequence-to-sequence problem, we propose a model named M4C-COO.
Fig.~\ref{fig:M4C-COO} illustrates M4C-COO.
M4C-COO is a variant of a model called multimodal multi-copy mesh (M4C)~\cite{M4C}.
M4C has been used for visual question answering with text (TextVQA~\cite{TextVQA,TextOCR}). 
M4C takes question word embedding, object, and OCR (text) tokens and fuses them using a multimodal transformer~\cite{transformer}.
In addition, M4C uses an iterative answer prediction mechanism to generate a multi-word answer.
Based on fused features and iterative answer prediction, M4C predicts the answer of the question.
Unlike M4C for TextVQA~\cite{M4C}, M4C-COO does not use question word embedding and object part. 
M4C-COO takes only onomatopoeia tokens and predicts a sequence of truncated texts.

To find truncated texts, we should exploit both visual and semantic (text) features because 1) truncated texts look similar and are close, and 2) They represent the intended meaning if they are connected.
M4C-COO exploits both visual and semantic features of onomatopoeias.
In M4C-COO, onomatopoeia tokens are embedded into four features, which are categorized as visual and semantic features.
For visual features, we use (1) appearance feature (FRCN) from onomatopoeia regions extracted by using Faster RCNN~\cite{fasterRCNN} part in MTS~v3~\cite{MTSv3}, and (2) 4-dimensional relative bounding box coordinates (bbox) for each onomatopoeia region.
For semantic features, we use (3) fastText~\cite{fastText} and (4) pyramidal histogram of characters (PHOC)~\cite{PHOC}.
fastText is a word embedding method with sub-word information.
fastText is well known for handling out-of-vocabulary words.
PHOC counts characters in the word and makes the pyramidal histograms for each word.

Furthermore, the copy mechanism of the pointer network~\cite{PointerNetwork} in M4C-COO is exactly what we needed for link prediction.
Generally, the copy mechanism selects a token (word) in the input sequence, and the selected token is used as an output token.
In other words, it copies the token in the input sequence to the output sequence.
In our task, we need to select truncated texts in the input onomatopoeia sequence.
This operation is the same as the copy mechanism.

M4C generally uses both thousands of vocabularies and copy mechanism to predict the output sequence. 
Thousands of vocabularies are used to generate a token that is not in the input sequence.
In our task, all the tokens in the output sentence are in input sentence, except for the delimiter symbol $<$d$>$ and end of sentence token $<$eos$>$.
Thus, \textit{M4C-COO uses only five vocabularies}: the delimiter symbol $<$d$>$ and four special tokens for training M4C-COO, padding token $<$pad$>$, start and end of sentence tokens ($<$sos$>$ and $<$eos$>$), and unknown token $<$unk$>$.

\section{Experiment and Analysis}\label{sec:experiment}
In this section, we present the results of the experiments on three tasks. 
Through experiments, we analyze the characteristics of COO and the limitations of the current methods.
More details of the experimental settings are provided in the supplementary materials.

\subsection{Implementation Detail} 
\PAR{Model and training strategy}
For text detection, we use the official codes of ABCNet~v2\footnote{https://github.com/aim-uofa/AdelaiDet/tree/master/configs/BAText}~\cite{ABCNetv2} and MTS~v3\footnote{https://github.com/MhLiao/MaskTextSpotterV3}~\cite{MTSv3}.
For text recognition and link prediction, we use the official codes of TRBA\footnote{https://github.com/clovaai/deep-text-recognition-benchmark}~\cite{TRBA} and M4C\footnote{https://github.com/facebookresearch/mmf/tree/main/projects/m4c}~\cite{M4C}, respectively.
For the training strategy, we follow the default setting to the extent possible.

\PAR{Dataset}
We split 109 comics of Manga109 into 89, 10, and 10 books and use them as training, validation, and test sets, respectively.
For the evaluation, we select the model with the best score on the validation set.
In each task, we use ground truth information rather than predicted results of other tasks.

\PAR{Evaluation metric}
For text detection, we use intersection over union to determine whether the model correctly detects the region of onomatopoeia.
For text detection and link prediction, we show precision (P), recall (R), and their harmonic mean (H, Hmean). 
As a default, we mainly use Hmean for comparison.
For text recognition, we show word-level accuracy for comparison.
We run three trials for all experiments and report average values.

\subsection{Text Detection}
We compare the effectiveness of bounding box annotation and polygon annotation, and compare the regression-based detector with the segmentation-based detector.

\PAR{Training with bounding box vs. with polygon}
Lines 1 and 3 in Table~\ref{tab:det} show the results of training with bounding box (the axis-aligned rectangle that bounds the onomatopoeia region) annotation instead of using polygon annotation.
Comparing lines (1 vs. 2) and (3 vs. 4), training with polygon annotation shows better performance than training with bounding box annotation: +4.8\% for ABCNet~v2 and +5.3\% for MTS~v3.
However, the performance improvement by using polygon annotation is not that considerable.
This result may indicate that current detection algorithms may not fully exploit polygon annotation.
To improve performance, we may need the algorithm that exploits irregular polygon annotations, such as partially shrunk polygon, more effectively.

\PAR{Regression vs. segmentation}
ABCNet~v2 and MTS~v3 are the representatives of regression-based and segmentation-based methods, respectively.
Comparing lines 2 and 4 in Table~\ref{tab:det}, MTS~v3 shows better performance +1.8\% than ABCNet~v2.
This result is unexpected and interesting because ABCNet~v2 shows better performance than MTS~v3 in other benchmark datasets such as MSRA-TD500~\cite{MSRA-TD500} (85.2 vs. 83.5).
This result indicates that segmentation-based methods can be advantageous for detecting the region of onomatopoeia.

\begin{table}[t] 
  \caption{Ablation study on text detectors ABCNet~v2 and MTS~v3}
  \vspace{-6mm}
  \tabcolsep=0.13cm
    \begin{center}
        \begin{tabular}{@{}llrrrrrr@{}}
            \toprule
            \# & Method & \multicolumn{1}{c}{P} & \multicolumn{1}{c}{R} & \multicolumn{1}{c}{H} \\ 
            \midrule
            1 & ABCNet~v2~\cite{ABCNetv2}-Bounding box & 61.9 & 60.7 & 61.2\\ 
            2 & ABCNet~v2~\cite{ABCNetv2}-Polygon & 67.7 & 64.5 & 66.0\\
            \midrule 
            3 & MTS~v3~\cite{MTSv3}-Bounding box & 67.5 & 58.2 & 62.5\\
            4 & MTS~v3~\cite{MTSv3}-Polygon & \textbf{69.8} & \textbf{65.9} & \textbf{67.8}\\
            \bottomrule
        \end{tabular}
    \label{tab:det}
    \end{center}
    \vspace{-2mm}
\end{table}

\begin{figure}[t]
\centering
    \includegraphics[width=\linewidth]{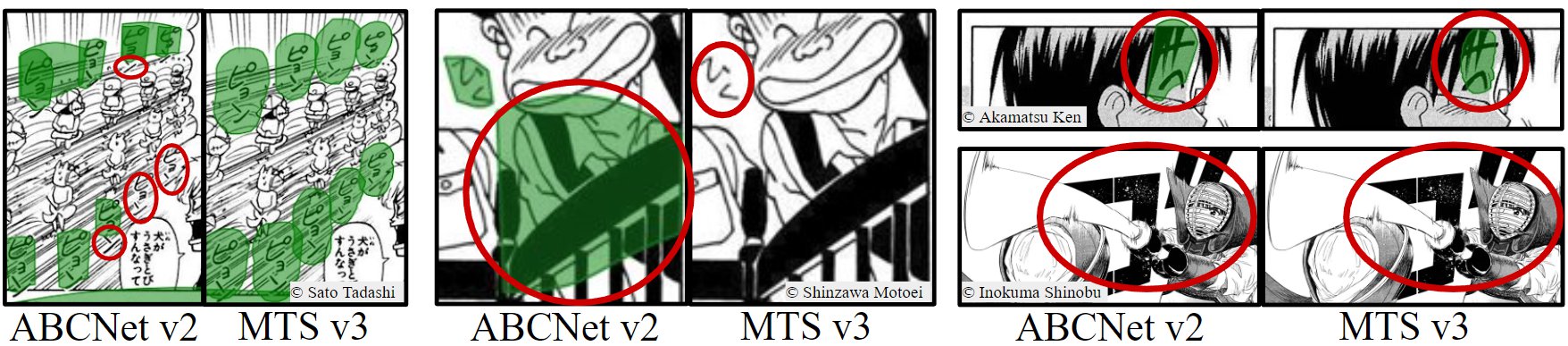}
    \vspace{-7mm}
    \caption{Text detection on the test set. The green regions are the predicted regions and the red circles are failures}
\label{fig:det}
\vspace{-2mm}
\end{figure}

\PAR{Visualization and failure case} Fig.~\ref{fig:det} shows the predictions on the test set and failure cases of the current methods.
Fig.~\ref{fig:det}~(left) shows that MTS~v3 correctly detects vertically long onomatopoeias (nine of ``ピョン''), whereas ABCNet~v2 misses two onomatopoeias (``ピョン'') and the part of onomatopoeias (``ン'').
According to MTS~v3 paper~\cite{MTSv3}, MTS~v3 is good at detecting rotated texts.
These cases show that MTS~v3 can also be used for vertically long texts.

Fig.~\ref{fig:det} shows other failure cases also:
(middle) shows MTS~v3 sometimes misses small onomatopoeias (the size of onomatopoeia ``ひく'' is $[width \times height] = [25 \times 28]$ whereas the image size is $[1654 \times 1170]$) and ABCNet~v2 misclassifies the text-like part (similar to ``ン'') as onomatopoeia. 
(right-top) shows that both methods misclassify the text-like part (similar to ``サへ'') as onomatopoeia. 
(right-bottom) shows that both methods miss the occluded texts. 

\begin{table}[t] 
\caption{Ablation study on text recognizer TRBA}
\vspace{-6mm}
  \tabcolsep=0.13cm
    \begin{center}
        \begin{tabular}{@{}llrr@{}}
            \toprule
            \# & Method & Accuracy \\ 
            \midrule
            1 & TRBA~\cite{TRBA} & 46.3 \\
            2 & + Rotation trick & 49.8 \\
            3 & + SAR decoding & 43.2 \\
            4 & + Rotation trick + SAR decoding & 55.4 \\
            \midrule
            5 & \#4 + Hard RoI (batch 100\%) & 63.5 \\
            6 & \#4 + Hard RoI (batch 50\%) & 67.9 \\
            \midrule
            7 & \#6 + 2D attention (height~64) & 78.5 \\
            8 & \#6 + 2D attention (height~100) & \textbf{81.0} \\
            \bottomrule
        \end{tabular}
    \label{tab:rec}
    \end{center}
    \vspace{-4mm}
\end{table}

\subsection{Text Recognition}
We investigate techniques to address vertically long texts, such as rotation and decoding tricks, Hard RoI masking, and 2D attention.

\PAR{Rotation and decoding tricks}
A text recognizer generally takes a text image in which characters are arranged horizontally as input and recognizes each character from left to right.
However, if the model takes a text image in which characters are arranged vertically as input, the model cannot recognize each character from left to right.
For this case, a simple rotation trick can be useful: Rotates vertical images 90 degrees to make them horizontal.
Here, an image whose height is greater than the width and whose text label contains more than two characters is regarded as a vertical image.
Comparing lines 1 to 2 in Table~\ref{tab:rec}, using the rotation trick results in a performance gain of +3.5\%.

Some cases, such as short vertical texts, are correctly recognized without the rotation trick but incorrectly recognized with the rotation trick.
For these cases, SAR decoding~\cite{SAR} can be useful.
SAR decoding is a decoding trick of the text recognition model SAR~\cite{SAR}: 
At the test time, if the height of the input image is greater than the width, 
the model takes three images as input data: the original image and images rotated by -90 and 90 degrees.
The confidence score on recognizing each image is calculated.
Next, the model outputs the result with the highest confidence score.
Lines 3 and 4 in Table~\ref{tab:rec} show that solely adding SAR decoding results in a performance drop of -3.1\%, whereas using both the rotation trick and SAR decoding improves the original TRBA by +9.1\%.

\begin{figure}[t]
\centering
    \includegraphics[width=0.9\linewidth]{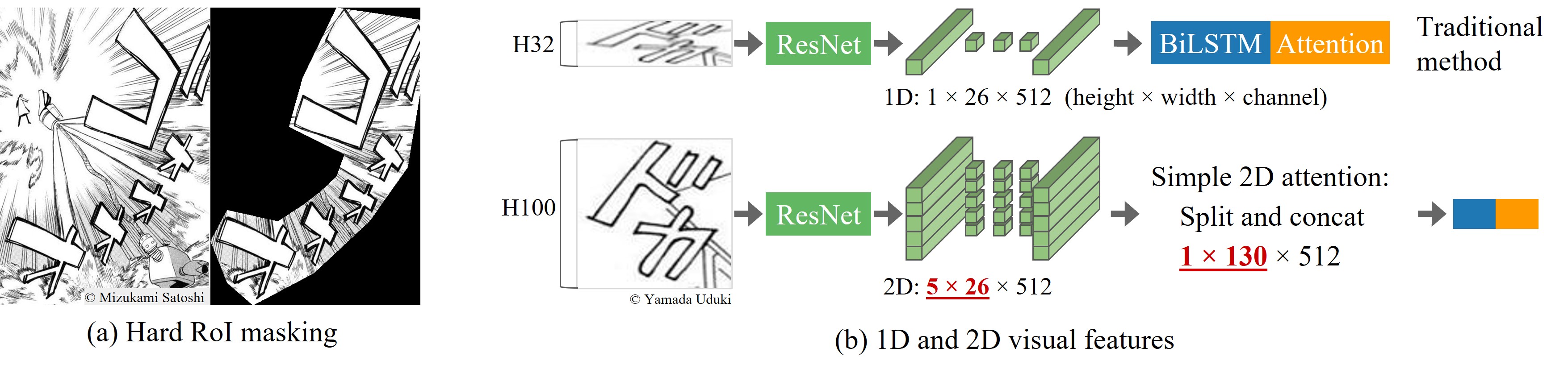} 
\vspace{-5mm}
\caption{(\textit{a}) Hard RoI masking removes the region irrelevant to text. 
(\textit{b}) Traditional methods use 1D attention on 1D visual features whereas 2D attention exploits 2D visual features}
\label{fig:rec}
\end{figure}
\vspace{2mm}

\PAR{Hard RoI masking}
When the text is diagonally long, the image contains more background noise, as shown in Fig.~\ref{fig:rec}~(a)~(left).
Background noise may be irrelevant to text and decrease performance.
To suppress this region, we use the hard region of interest (Hard RoI) masking~\cite{MTSv3} that removes this region, as shown in Fig.~\ref{fig:rec}~(a)~(right).
Comparing lines 4 and 5 in Table~\ref{tab:rec}, using Hard RoI masking shows an improvement by +8.1\%.
Furthermore, considering that the evaluation is conducted without Hard RoI masking, teaching the model how to handle the original images can be useful.
To do so, we fill half of each mini-batch with the original images.
As shown in line 6, the performance further improves by +4.4\%.

\begin{figure}[t]
\centering
     \includegraphics[width=0.9\linewidth]{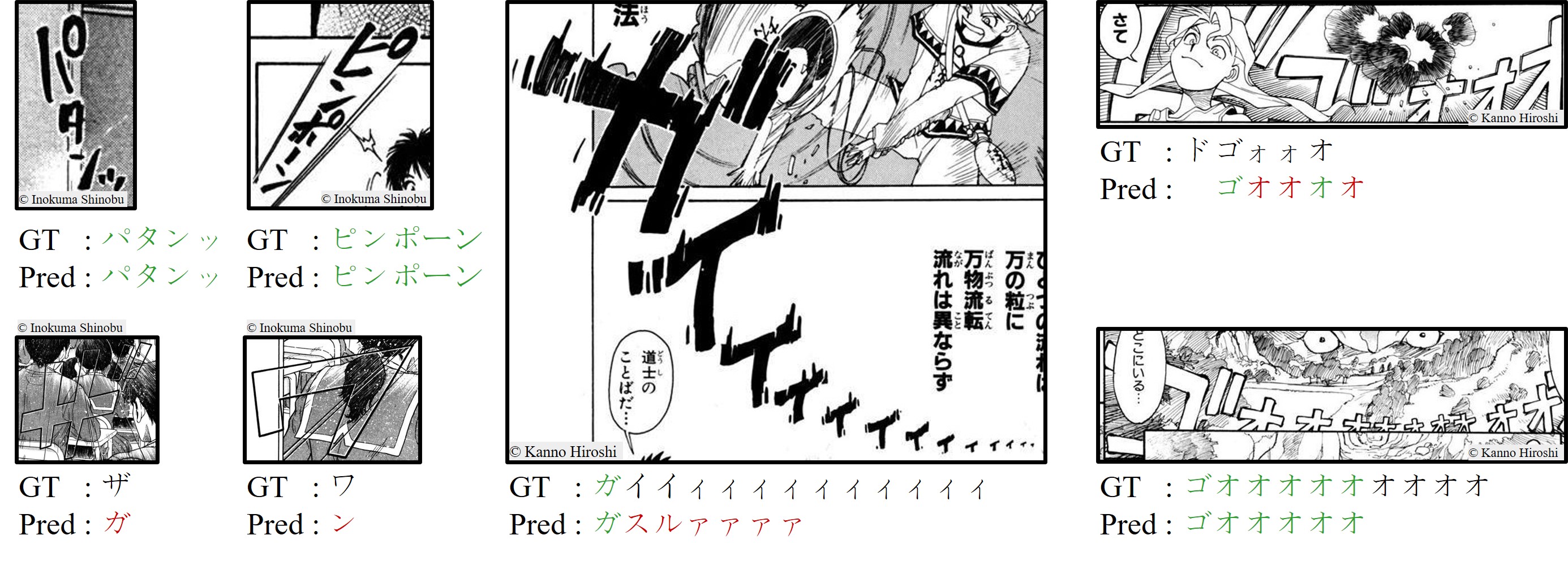}
     \vspace{-5mm}
    \caption{Text recognition on the test set. 
    GT denotes the ground truth, and Pred denotes the prediction by TRBA with the best score (\#8 in Table~\ref{tab:rec}).
    Green- and red-colored characters denote correct and incorrect recognition, respectively}
\label{fig:vis-rec}
\end{figure}

\PAR{2D attention on 2D visual features}
The model can benefit from considering the attention on vertical direction.
Traditional methods~\cite{CRNN,RARE,FAN,ASTER,TRBA} take an image with a height of 32 and make 1D visual features through convolutional networks (ResNet), as shown in Fig.~\ref{fig:rec}~(b)~(top): $[height \times width \times channel]=[1 \times 26 \times 512]$ is called 1D because the vertical dimension is squeezed to 1.
Because English texts are mainly horizontal, most methods follow this trend.
Recently, some methods~\cite{ATR,SAR,EPAN,satrn,MASTER} take an image whose height is greater than 32 and make 2D visual features, as shown in Fig.~\ref{fig:rec}~(b)~(bottom): $[\textit{5} \times 26 \times 512]$ is called 2D because the vertical dimension remains at \textit{5}.
These methods showed performance improvements by using 2D attention on 2D visual features. 

We show the effectiveness of 2D attention with a minimal modification of the model.
We use a simple 2D attention method to exploit 2D visual features, as shown in Fig.~\ref{fig:rec}~(b)~(bottom).
We use 1D BiLSTM, and we need to make vertical dimension 1 before BiLSTM.
To do so, we split vertical features and concatenate them horizontally (e.g., $5 \times 26$ to $1 \times 130$), as done in EPAN~\cite{EPAN}.
Lines 7 and 8 in Table~\ref{tab:rec} show the performance improvements.
Simple 2D attention improves performance by +10.6\% and +13.1\% where the heights of input images are 64 and 100, respectively.

\PAR{Visualization and failure case}
Some vertical or diagonal texts can be correctly recognized, whereas transparent, partially shrunk, or occluded texts are incorrectly recognized.
Fig.~\ref{fig:vis-rec} shows the predictions by TRBA with the best score (\#8 in Table~\ref{tab:rec}):
TRBA correctly recognizes vertically long texts ``パタンッ'' and ``ピンポ---ン'' (row 1, column 1 and 2) whereas incorrectly recognizes transparent texts ``ザ'' and ``ワ'' (row 2, column 1 and 2), partially shrunk text (column 3), and texts occluded by objects or a frame (column 4).

\begin{table}[t] 
\caption{Ablation study on link prediction model M4C-COO}
\vspace{-6mm}
  \tabcolsep=0.13cm
    \begin{center}
        \begin{tabular}{@{}llllrrr@{}}
            \toprule
            \# & Method & Visual feature & Semantic feature & \multicolumn{1}{c}{P} & \multicolumn{1}{c}{R} & \multicolumn{1}{c}{H} \\ 
            \midrule
            1 & Rule-base & distance & $-$ & 1.1 & \textbf{74.5} & 2.1\\ 
            \midrule
            2 & M4C-COO & FRCN + bbox & fastText + PHOC & \textbf{77.2} & 68.7 & \textbf{72.7} \\
            3 & + Vocab. 11,640 & FRCN + bbox & fastText + PHOC & 55.0 & 38.7 & 45.4 \\
            \midrule
            4 & Only fastText & $-$ & fastText & 42.4 & 30.2 & 35.2 \\
            5 & + PHOC & $-$ & fastText + PHOC & 61.7 & 50.5 & 55.4 \\
            6 & + PHOC + FRCN  & FRCN & fastText + PHOC & 62.1 & 53.4 & 57.2 \\
            \bottomrule
        \end{tabular}
    \label{tab:link}
    \end{center}
    \vspace{-4mm}
\end{table}

\subsection{Link Prediction}
We present the results of the link prediction task, such as a comparison with a rule-based method and ablation studies.

\PAR{M4C-COO vs. distance-based method}
We test the distance-based method as a baseline:
1) Calculate the average distance from one truncated text to another truncated text.
2) For each onomatopoeia, if the other onomatopoeia is closer than the average distance, they are regarded as linked.
Line 1 in Table~\ref{tab:link} shows that the distance-based method has the highest recall of 74.5\% but a considerably low precision of 1.1\%.
Comparing lines 1 and 2, M4C-COO shows a much better performance of +70.6\% (Hmean) than the distance-based method.

\PAR{Effect of vocabulary}
Comparing lines 2 and 3 in Table~\ref{tab:link}, 
\textit{M4C-COO (with only five vocabularies) shows a much better performance of +27.3\% than M4C-COO with 11,640 vocabularies (vocabularies of the training set)}.
This result indicates that using only the copy mechanism is more suitable for this task than using both many vocabularies and the copy mechanism.
Many vocabularies may disturb the copy mechanism, and therefore performance decreases.

\PAR{Ablation study}
Line 4 in Table~\ref{tab:link} shows that using only fastText feature for M4C-COO results in a performance drop of -37.5\%.
Line 5 shows that if we use PHOC together, the performance drastically improves by +20.2\%.
Line 6 indicates that if FRCN (appearance feature) is added, the performance further improves by +1.8\%. 
However, the performance gain by adding FRCN is considerably less than that by adding PHOC.
This result is reasonable considering the drawing style.
Because the comic artist is identical in each image, the drawing (writing) style of onomatopoeias in each image is similar.
Therefore, exploiting the appearance feature to predict the link is less effective.

Comparing lines (2 vs. 6), using bbox (relative coordinate information) makes huge improvement by +15.5\%.
Sometimes there are multiple onomatopoeias whose texts are identical in an image, and only one of them is a truncated text linked to other truncated text.
In such cases, the model can benefit from bbox. 
The model can select the only one truncated text by using coordinate information.

\begin{figure}[t]
\centering
     \includegraphics[width=0.9\linewidth]{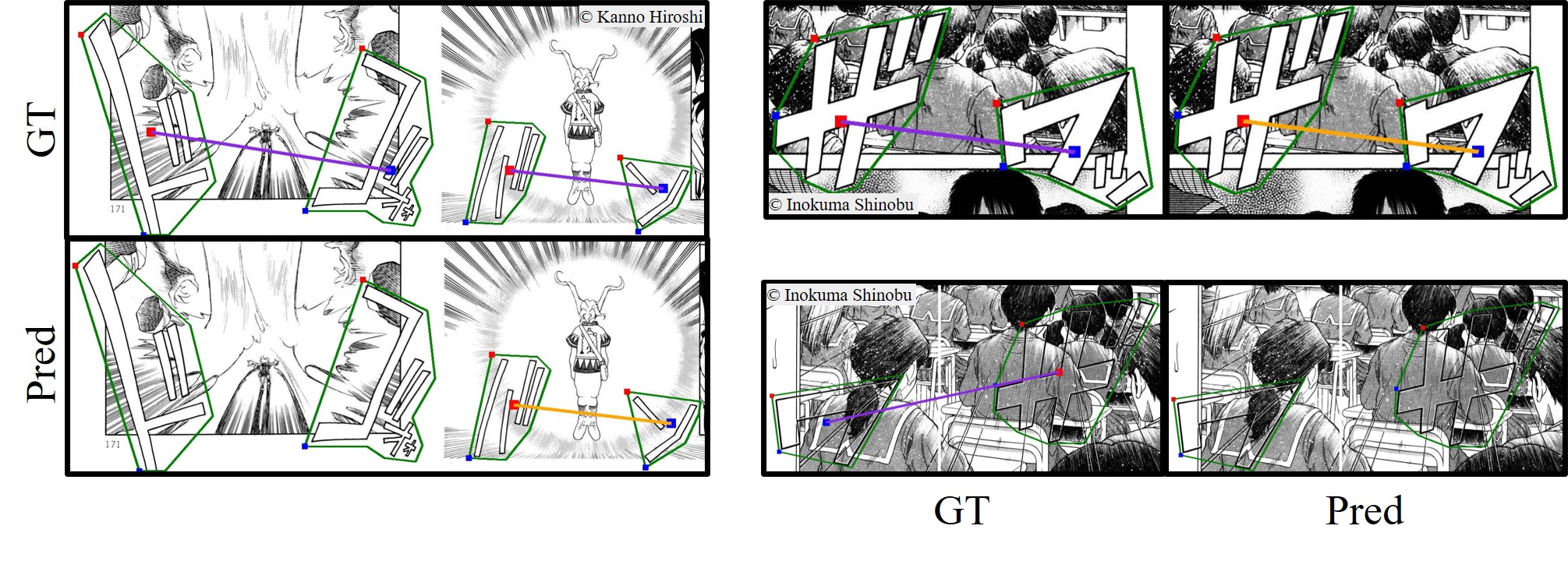}
     \vspace{-6mm}
    \caption{Link prediction on the test set. 
    The purple line denotes the ground truth link between truncated texts, and the orange line denotes the predicted link}
\label{fig:vis-link}
    \vspace{-2mm}
\end{figure}

\PAR{Visualization and failure case}
Some links between truncated texts can be correctly predicted, whereas predicting links between texts similar to background images is difficult.
Fig.~\ref{fig:vis-link} shows the predictions by M4C-COO:
(column 1) shows that M4C-COO correctly predicts the link between ``バ'' and ``ン'', while misses the link between ``ド'' and ``ゴォォ''.
M4C-COO correctly predicts the link between ``ザ'' and ``ワッ'' (column 2, row 1) and misses the link between transparent texts ``ザ'' and ``ワ'' (column 2, row 2).

\section{Conclusion}\label{sec:conclusion}
We have constructed a novel dataset named COO.
COO contains many arbitrary-shaped texts or arbitrarily placed texts.
Some texts are separated into several parts, and each part is a truncated text.
To capture the intended meaning of truncated texts, we have proposed the link prediction task and the M4C-COO model.
We have conducted three tasks (text detection, text recognition, and link prediction) and provided decent baselines.
We have experimentally analyzed the characteristics of COO and the limitation of current methods.
Detecting the onomatopoeia region and capturing the intended meaning of truncated texts are not straightforward.
Thus, COO is a challenging text dataset.
We hope that this work will encourage future work on recognizing various types of texts.

\section*{Acknowledgements}
This work is partially supported by JSPS KAKENHI
Grant Number 22J13427, JST-Mirai JPMJMI21H1 and AI center of the Univ. of Tokyo.

\clearpage
\appendix

\section*{Supplementary Material}
The following materials are provided in the supplementary material:

\noindent Supplement~\ref{sup:data}: We show more details of the visualization, annotation guideline, statistics, and analysis of our dataset COO.

\noindent Supplement~\ref{sup:method}: We describe the details of text detection and link prediction methods in \S\ref{sec:method}.

\noindent Supplement~\ref{sup:experiment}: We provide more details of experimental settings.

\section{More Details of Comic Onomatopoeia Dataset}\label{sup:data}
We have created the dataset COO to encourage studies on more irregular texts and to further improve text detection, recognition, and link prediction methods.
Recognizing Japanese comic onomatopoeias is very difficult. 
If a model can recognize them, we expect that the model can also recognize other less difficult texts.
A similar case exists in another task. 
Although Manga109, the base of COO, is a Japanese comic dataset, Manga109~\cite{manga109} is widely used as a benchmark dataset in the super-resolution task~\cite{Zhang_2018_ECCV,niu2020single}.
If a model works not only in the major domain but also in minor sub-domain such as Japanese comics, we expect that the model has the potential to be generalized to various minor sub-domains.

\subsection{Details of Annotation Guidelines}
We define comic onomatopoeia as the texts that represent the sound or state of objects.
In comics, the dialogue (line or quote) is usually in speech balloons.
However, the dialogue is sometimes outside of speech balloons and written in similar fonts to onomatopoeias. 
For example, when the character shouts another character's name, the name is sometimes written in informal fonts.
The dialogue written in informal fonts confused the annotators whether it should be regarded as an onomatopoeia.
Following our guideline, they are not regarded as onomatopoeias because they do not represent the sound or state of objects.

When the annotators encountered ambiguous cases such as the dialogue written in informal fonts, 
annotation checker and we (authors) discussed how to handle them.
To determine whether the text is onomatopoeia or not, we followed two main rules:
1) If the text is the part of the dialogue or similar to the dialogue, the text is not regarded as onomatopoeia.
2) If the text is not the part of the dialogue and represents the sound or state of objects, the text is regarded as onomatopoeia.
In addition, we take into account that the onomatopoeias are usually written in informal fonts.

For the link annotation, we take into account the order of reading truncated texts.
For example, if the word ``ばさ'' is separated into ``ば'' and ``さ'', we annotate ``ば'' then ``さ'' rather than ``さ'' then ``ば''.

\subsection{Data Preprocessing} 
In Japanese comics, there were also a few English onomatopoeias. 
The number of English onomatopoeias was only 148.
They were excluded because they were not matched with our guideline. 
Most English onomatopoeias were verbs and did not represent the sound or state of objects.

All polygons in COO were validated by the function of \texttt{object.is\_valid} in the python library shapely\footnote{https://shapely.readthedocs.io/en/stable/manual.html}~\cite{shapely}.
Polygons that have unexpected intersections therein were corrected.

Based on the comic artist of each comic, we split 109 books in Manga109~\cite{manga109} into training, validation, and test sets.
In Manga109, there are multiple books written by the same comic artist.
We split them into training and test sets.
For example, the books ``LoveHina\_vol1'' and ``LoveHina\_vol14'' are written by the comic artist Akamatsu Ken.
The books ``ByebyeC-BOY'' and ``TotteokiNoABC'' are written by the comic artist Aida Mayumi.
Overall, Manga109 contains 30 books written by 15 comic artists (2 books per comic artist), 3 books written by one comic artist, and 4 books written by another comic artist.

\begin{table*}[t] 
\caption{The intended meaning after connecting truncated texts based on each link. 
    Each example is in Fig.~\ref{fig:data}~(b). 
    Row and column denote the row and column of Fig.~\ref{fig:data}~(b), respectively. 
    Text 1 and Text 2 are truncated texts.}
\vspace{-6mm}
  \tabcolsep=0.095cm
    \begin{center}
        \begin{tabular}{@{}lllll@{}}
            \toprule
            (Row, Column) & Text~1 & Text~2 & Connected text & Intended meaning of connected text\\
            \midrule
            (1, 1) & く & らっ & くらっ & State of feeling dizzy\\
            (1, 2) & ガ--- & ---ッ & ガ--- ---ッ & State of moving forward vigorously \\
            (2, 1) & ガシャ & $\cdot\cdot$ン & ガシャ$\cdot\cdot$ン & Sound of breaking something \\
            (2, 2) & と & ん & とん &  Sound of putting something down\\
            \bottomrule
        \end{tabular}
    \label{tab:connect}
    \end{center}
\end{table*}

\subsection{Intended Meaning of Truncated Texts}
To correctly capture intended meaning of truncated texts, we predict the link between truncated texts.
With the predicted link, we connect truncated texts and capture the intended meaning.
Table~\ref{tab:connect} shows the meaning of connected text after connecting truncated texts based on the link: each example is illustrated in Fig.~\ref{fig:data}~(b) of the main text.

Some truncated texts contain special characters such as ``!'', ``?'', ``～'', and so on.
We consider that they are also needed to capture the intended meaning, and they have a link with other truncated text. 

\begin{figure}[t]
    \centering
    \includegraphics[width=0.7\linewidth]{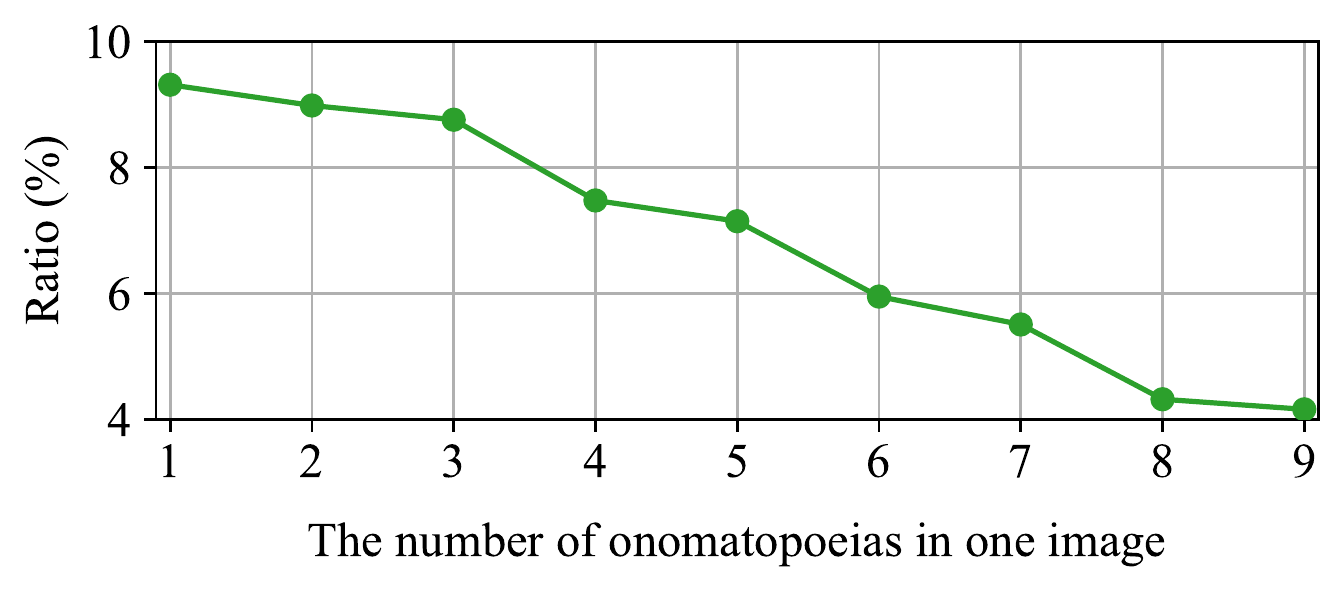}
    \vspace{-4mm}
    \caption{The ratio of the number of onomatopoeias in one image}
    \label{fig:ratio}
\end{figure}

\begin{table}[t]
\tabcolsep=0.14cm
\caption{More statistics of COO dataset: (a) Top 10 vocabularies in COO, sorted by the frequency, (b) The number of onomatopoeia according to its length of text}
\centering
\begin{tabular}{ccc}
(a) Top 10 vocabularies in COO & & (b) Count by the length of text \\[1mm] 
    \begin{tabular}{@{}lrr@{}}
        \toprule
        \multicolumn{1}{l}{Vocabulary} & \multicolumn{1}{l}{Count} & \multicolumn{1}{l}{Ratio (\%)} \\
        \midrule
        オ & 891 & 1.4 \\
        ザワ & 749 & 1.2 \\
        パチ & 687 & 1.1 \\
        ワー & 677 & 1.1 \\
        ドキ & 365 & 0.6 \\
        ン & 347 & 0.6 \\
        はっ & 346 & 0.6 \\
        ? & 333 & 0.5 \\
        ゴ & 309 & 0.5 \\
        はあ & 308 & 0.5 \\
        \bottomrule
    \end{tabular} & \qquad \qquad \qquad &
    
    \begin{tabular}{@{}crr@{}}
        \toprule
        \multicolumn{1}{l}{Length of text} & \multicolumn{1}{l}{Count} & \multicolumn{1}{l}{Ratio (\%)} \\
        \midrule
        1 & 5,845 & 9.5 \\
        2 & 26,064 & 42.4 \\
        3 & 15,909 & 25.9 \\
        4 & 6,713 & 10.9 \\
        5 & 3,384 & 5.5 \\
        6 & 2,018 & 3.3 \\
        7 & 747 & 1.2 \\
        8 & 410 & 0.7 \\
        9 & 183 & 0.3 \\
        10 & 75 & 0.1 \\
        \bottomrule
    \end{tabular} \\ 
\end{tabular} 
\label{tab:freq-length}
\end{table}

\subsection{More Statistics and Analysis}
The dataset COO can be used to translate Japanese comics or analyze Japanese comics or onomatopoeias.
In this subsection, we show more analysis of onomatopoeias in Japanese comics.

Fig.~\ref{fig:ratio} shows the ratio of the number of onomatopoeias in one image.
About 47.7\% images have more than four onomatopoeias.
About 18.3\% images have only one or two onomatopoeias.
About 17.8\% images have no onomatopoeias, and about 20.6\% images have more than nine onomatopoeias (both are not listed in the graph).
These results indicate that Japanese comic images usually contain many onomatopoeias.

\begin{figure}[t]
\centering
    \includegraphics[width=0.7\textwidth]{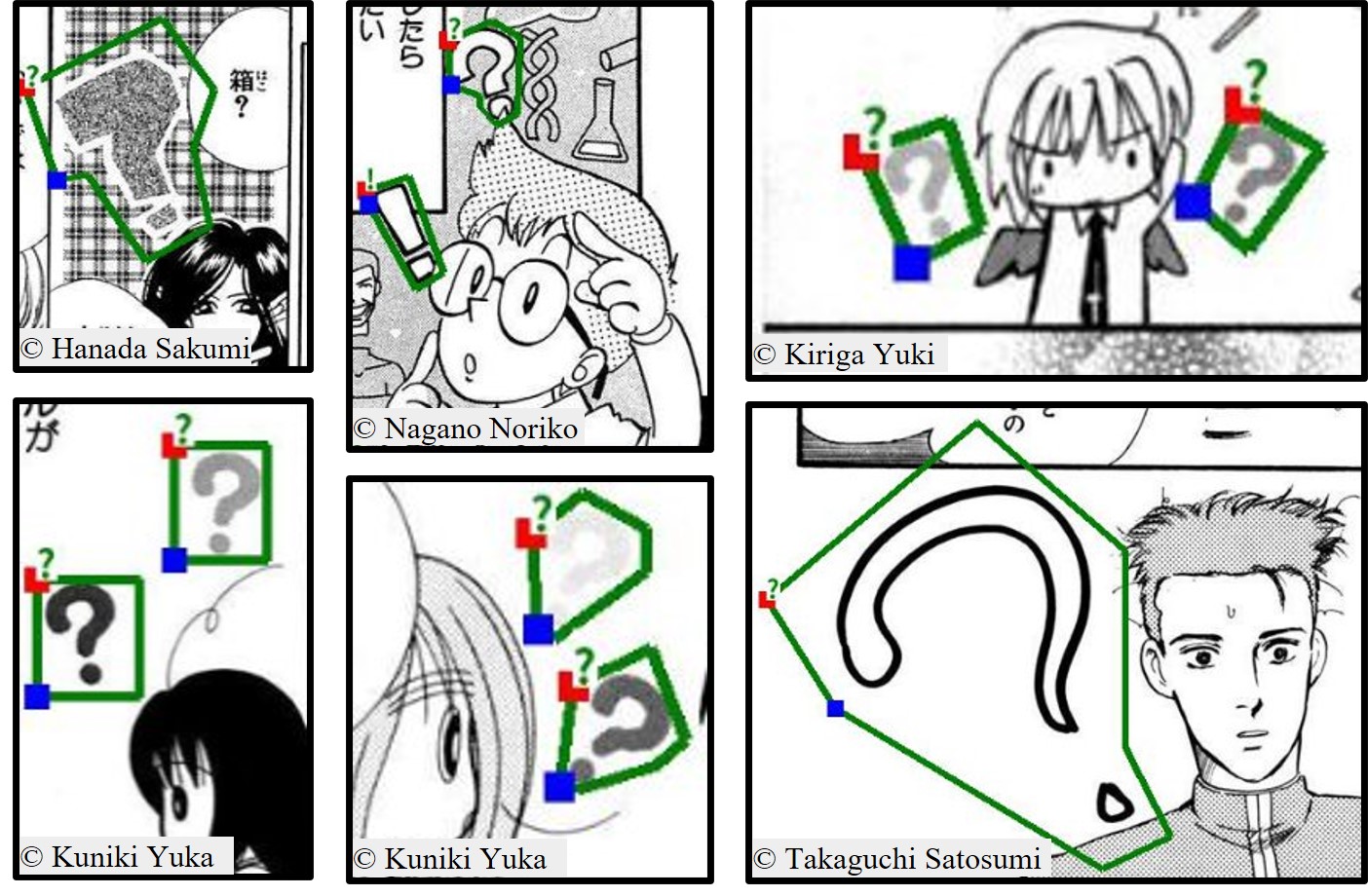}
    \caption{
    Question marks are also regarded as onomatopoeias because they represent the state where objects (human) wonder about something
    }
    \label{fig:question}
\end{figure}

Table~\ref{tab:freq-length}~(a) shows top 10 vocabularies sorted by the frequency.
Overall, all top 10 vocabularies are short (less than 3 characters) and the sound or state of people are frequently used.
``オ'' and ``ワ'' represent the sound of yelling by people. 
``ザワ'' and ``ゴ'' represent the state of the scene or atmosphere.
``パチ'' is the sound of applause by people.
``ドキ'' represents the state or sound of the heart beating.
``はっ'' represents the sound of noticing something.
``はあ'' represents the sound of a sigh.
In COO, question mark ``?'' is also regarded as onomatopoeia because they represents the state in which people wonder about something. 
``?'' is usually written in informal fonts as shown in Fig.~\ref{fig:question}.

Table~\ref{tab:freq-length}~(b) shows the number of onomatopoeias according to its length of text.
The length of most onomatopoeias is two (42.4\%) or three (25.9\%) characters.
It indicates that there are many short onomatopoeias in Japanese comics.
COO consists of many short onomatopoeias and some long onomatopoeias.
Even though many are short, it is still difficult to detect or recognize them because they are written in informal fonts, arbitrary-shaped, placed at unexpected position, or occluded, as shown in Fig.~\ref{fig:data2}, \ref{fig:data3}, and \ref{fig:data4}.

\subsection{More Visualization of COO} 
\begin{figure*}[t]
\centering
     \includegraphics[width=\linewidth]{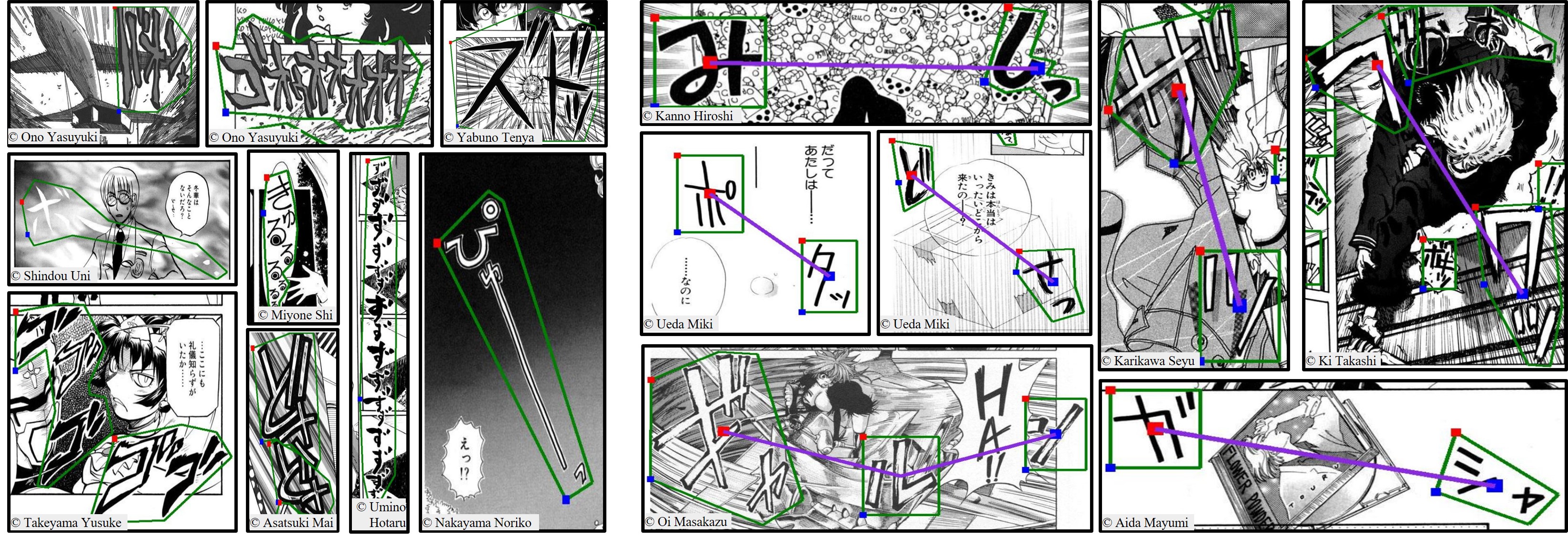}
    \vspace{-6mm}
    \caption{
    Each example shows diversity of onomatopoeias.
    Some are on the objects, some are written in informal fonts, and others lie across multiple frames in comics.
    Red and blue squares denote the start and end points of each annotation, respectively.
    Purple lines denote the link between truncated texts
    }
    \label{fig:data2}
    \vspace{-2mm}
\end{figure*}

\begin{figure*}[t]
\centering
     \includegraphics[width=\linewidth]{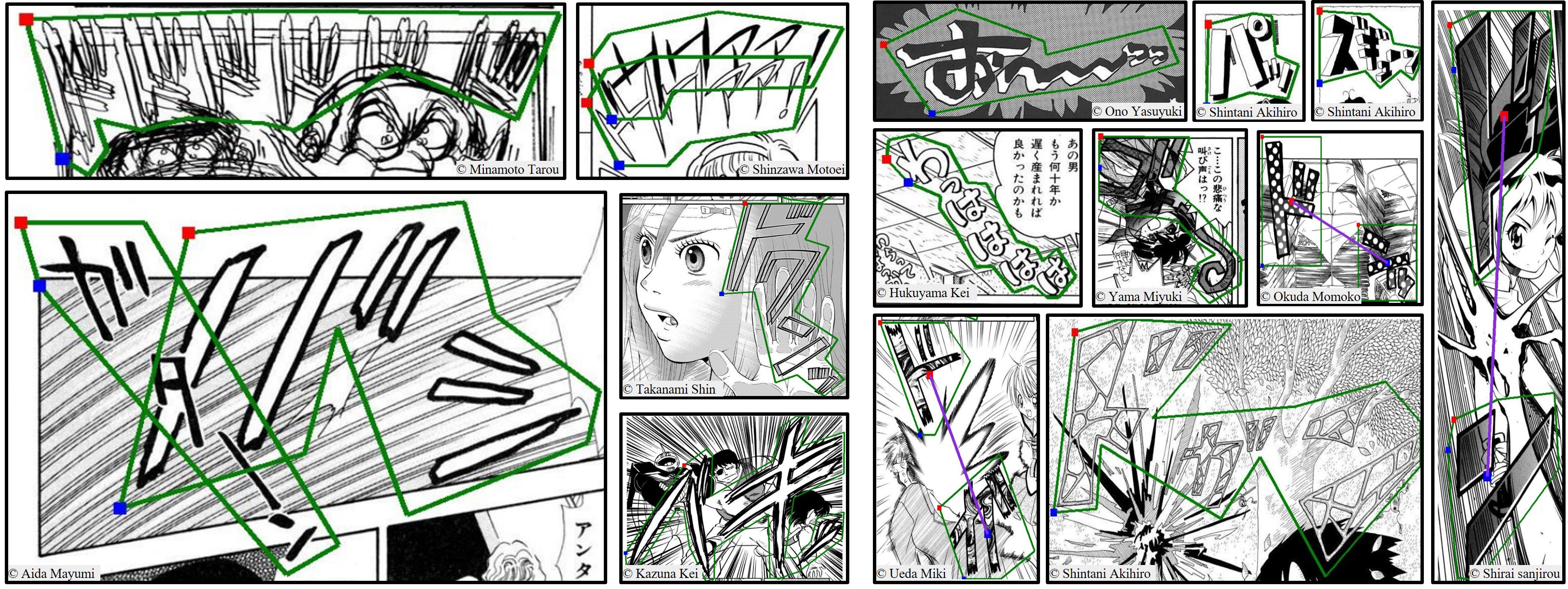}
    \vspace{-6mm}
    \caption{
    Each example shows diversity of onomatopoeias.
    Some are overlapped with other onomatopoeias and others are transparent texts that look similar to background objects
    }
    \label{fig:data3}
    \vspace{-2mm}
\end{figure*}

\begin{figure*}[t]
\centering
     \includegraphics[width=\linewidth]{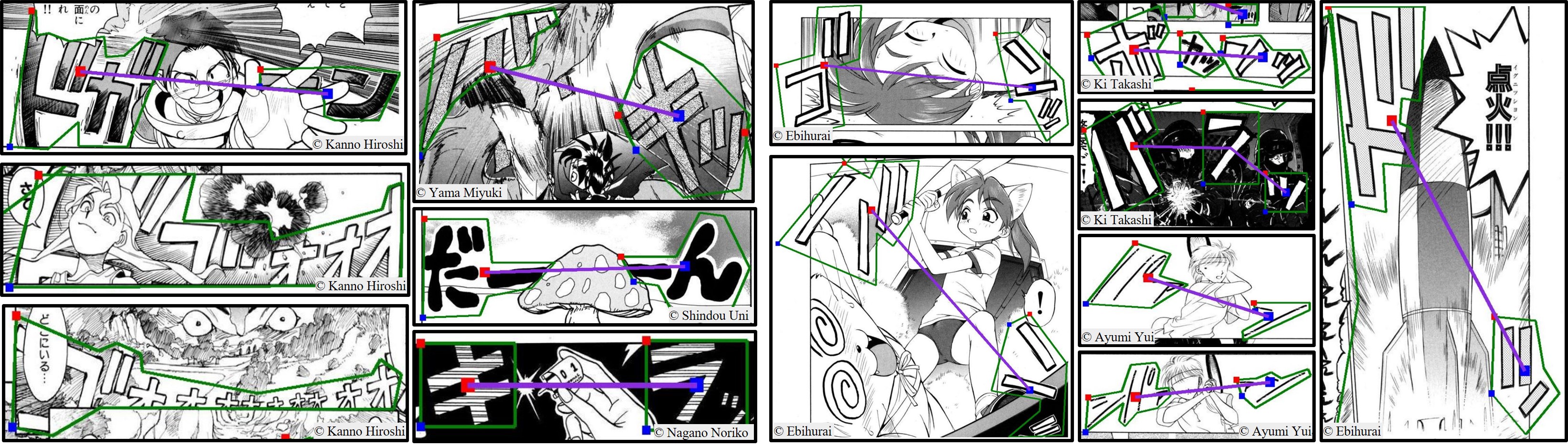}
    \vspace{-6mm}
    \caption{
    Each example shows diversity of onomatopoeias.
    Some are occluded by objects or frames, and others are separated into several parts
    }
    \label{fig:data4}
    \vspace{-2mm}
\end{figure*}

Fig.~\ref{fig:data2}, \ref{fig:data3}, and \ref{fig:data4} show various onomatopoeias, such as arbitrary texts and truncated texts with link annotations.
Some are transparent texts that look similar to background objects, some are overlapped with other onomatopoeias, and others are occluded by objects or frames.

\clearpage

\clearpage

\section{More Details of Methods}\label{sup:method}

\subsection{Text Detection}
As described in \S\ref{sec:method}, we use ABCNet~v2~\cite{ABCNetv2} and MTS~v3~\cite{MTSv3}. 
We use only the text detection part of them and use only the loss function corresponding to the text detector part.

Regression-based methods for text detection usually find some points that represent each text region.
ABCNet~v2 finds eight control points for each text region.
Eight control points represent two Bezier curves.
One Bezier curve draws the top line of the text region, and another Bezier curve draws the bottom line.
ABCNet~v2 is trained with the regression loss for finding the coordinate of eight control points.

For MTS~v3, the official code\footnote{https://github.com/MhLiao/MaskTextSpotterV3}~\cite{MTSv3} has the option for ``train detection only,'' and we used this option. 
In this option, the model does not use the heads on the region of interest (RoI heads) and the model is trained with only segmentation loss for each text region.

\subsection{Link Prediction}
For M4C-COO model, we use two visual features (FRCN and bbox) and two semantic features (fastText~\cite{fastText} and PHOC~\cite{PHOC}) 

\PAR{FRCN} It is the visual (appearance) feature extracted from Faster RCNN~\cite{fasterRCNN}. 
In M4C~\cite{M4C} model, the feature of the fc6 layer in Faster RCNN is used.
However, because we do not use RoI heads in MTS~v3, we cannot use the feature of the fc6 layer for our M4C-COO model.
Instead, we use the feature of the last layer, which is the segmentation map.
We use the segmentation map, expecting that the segmentation map suppresses the elements that are irrelevant to texts and grasps the shape of texts.
We pool the regions of proposals in the segmentation map into $32\times32$. 
After reshaping them into 1024 dimensions ($32\times32=1024$), we use them as FRCN.

\PAR{bbox} It is the visual (location) feature. 
Each bbox is 4-dimensional relative bounding box coordinates and is calculated as follows.
\begin{align}
bbox = \left [\frac{x_{min}}{W_{im}}, \frac{y_{min}}{H_{im}}, \frac{x_{max}}{W_{im}}, \frac{y_{max}}{H_{im}} \right ]
\end{align}
where $W_{im}$ and $H_{im}$ denote width and height of an input image, respectively.

\PAR{fastText} It is the semantic feature, a word embedding method that considers subword information.
In other words, when training the word embedding model, we also use the n-gram (subword) of each word as training data.
For example, if we use trigram as the subword information for the word ``comics'', we also use ``$<$co'', ``com'', ``omi'', ``mic'', ``ics'', and ``cs$>$'' as the training data.
$<$ and $>$ are boundary symbols that denote at the beginning and end of the word.
Because the fastText model is trained with subwords, fastText can handle out-of-vocabulary words if their subwords are used in training.

\clearpage

\begin{wrapfigure}{r}{0.5\textwidth}
\centering
    \includegraphics[width=0.5\textwidth]{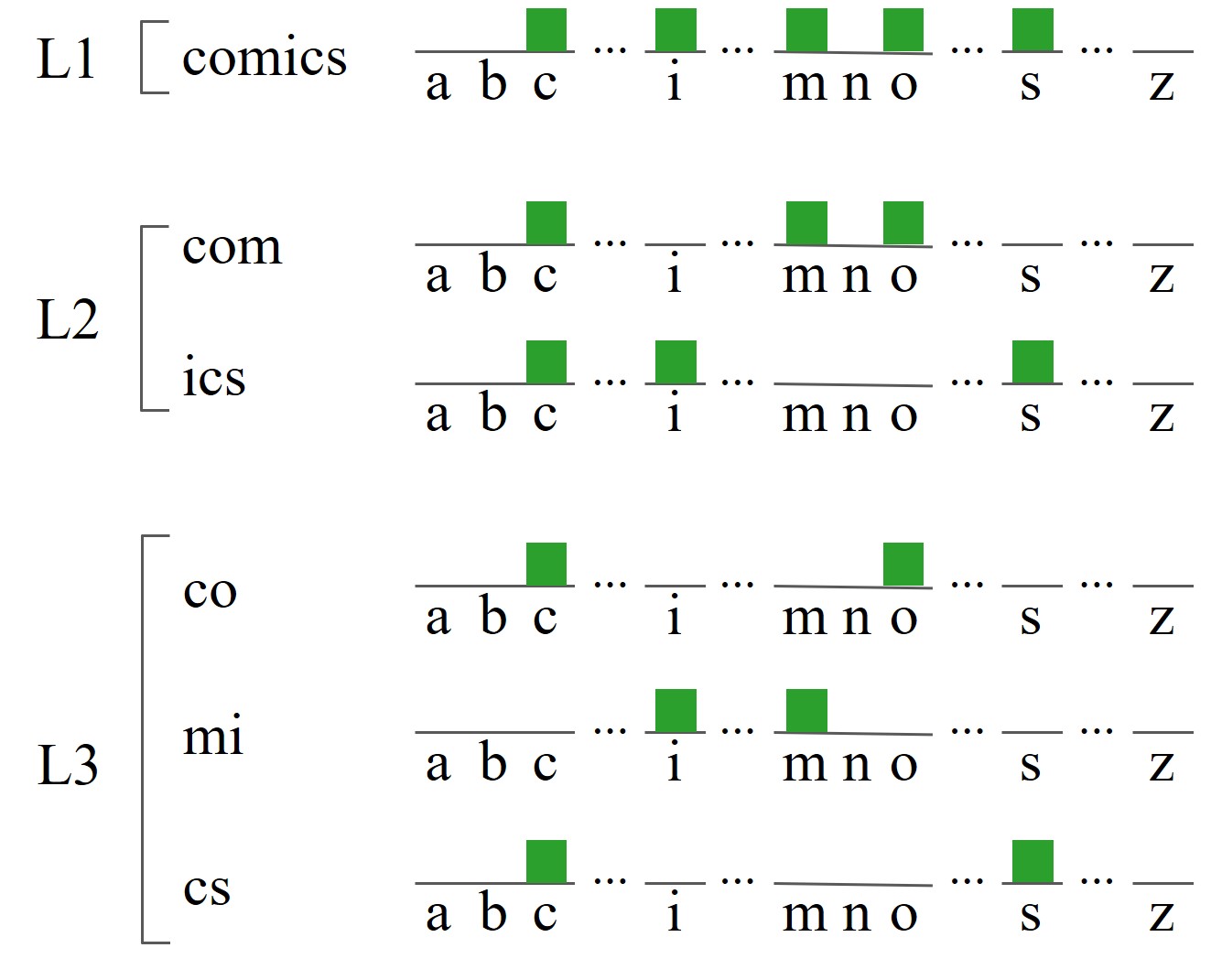}
    \caption{Pyramidal histogram of characters (PHOC). 
    (\textit{L1}), (\textit{L2}), and (\textit{L3}) are histograms of a word at levels 1, 2, and 3, respectively
    }
    \label{fig:PHOC}
    \vspace{-2mm}
\end{wrapfigure}

\PAR{PHOC} It is the semantic feature, the pyramidal histograms of characters (PHOC) for each word.
Fig.~\ref{fig:PHOC} illustrates the concept of PHOC for English characters.
PHOC is the concatenation of multiple binary histograms.
If we embed the word ``comics'' as a binary histogram of characters, we will get a histogram like as L1 in Fig.~\ref{fig:PHOC}.
Each dimension of the histogram represents whether the word ``comics'' contains a character or not.
However, this embedding method has a problem: words ``comics'' and ``cosmic'' share the same histogram.
To avoid this problem, the pyramid version of the histogram of characters is proposed by \cite{PHOC}.
PHOC counts characters in the part of the word instead of the whole word.
For example, at level 2, the word ``comics'' split into the first half of the word ``com'' and the second half of the word ``ics''.
After that, we construct 2 histograms for each half word, like as L2 in Fig.~\ref{fig:PHOC}.
At level 3, we split the word ``comics'' into three parts ``co'', ``mi'', and ``cs'', and then conduct the same thing.
The concatenation of these binary histograms is the PHOC representation.
In practice, levels 2, 3, 4, and 5 leading to a histogram of $(2+3+4+5)\times36=504$ 
(36 is the sum of 26 lower-case alphabets and 10 digits) are used for English characters.
In addition, the 50 most common English bigrams with level 2 leading to 100 dimensions ($2\times50=100)$ is also used. 
As a result, PHOC is a 604-dimensional histogram for English.
In COO, we use 182 characters for Japanese Hiragana, Katakana, and symbols.
As a result, PHOC is a 2648-dimensional histogram: $(2+3+4+5)\times182+2\times50=2648$.

\section{More Details of Experimental Setting}\label{sup:experiment}

\subsection{Text Detection}
For ABCNet~v2, two Bezier curves are created based on the points of each polygon region.
Then, we generate eight points that represent the two Bezier curves.
We use coordinates of these eight points as training data.
When the number of points of the onomatopoeia region is four, we should have interpolated additional two points in the top line and bottom line as the authors of ABCNet~v2~\cite{ABCNetv2} did. 
As a result, the top and bottom lines have three points, respectively.
When we did not interpolate, two Bezier curves were incorrectly created from four points.

To detect onomatopoeia regions, we conduct fine-tuning ABCNet~v2 and MTS~v3 on our dataset COO by using the pretrained models on the dataset CTW1500~\cite{CTW1500}.
With four NVIDIA Tesla V100 GPUs, training of ABCNet~v2 and MTS~v3 take about 21 and 33 hours, respectively.

\subsection{Text Recognition}
According to Baek~\etal~\cite{STRfewerlabels}, most text recognition methods have been trained on synthetic data because the number of real data was too small. 
They also shows that if we have enough real data, we can train a text recognizer only with real data.
In our case, we have enough data to train a text recognizer, and thus we did not use synthetic data.
Following Baek~\etal~\cite{STRfewerlabels}, we use Adam~\cite{adam} optimizer and an one-cycle learning rate schedule~\cite{super-convergence} for faster training and better performance.
With one NVIDIA Tesla V100 GPU, training of TRBA with heights of the input image 32, 64, and 100 takes about 13, 24, and 35 hours, respectively.

\subsection{Link Prediction}
For distance-based method, we calculate the average distance from one truncated text to another truncated text.
In the case of training data, the average distance is 266.2. 
For each feature of M4C-COO, the dimensions of FRCN, bbox, fastText, and PHOC are 1024, 4, 300, and 2648, respectively.
With one NVIDIA Tesla V100 GPU, training of M4C-COO takes about 2 hours.

%
%
\bibliographystyle{splncs04}
\bibliography{egbib}
\end{document}